\documentclass{article}
\usepackage{arxiv}

\usepackage[utf8]{inputenc} 
\usepackage[T1]{fontenc}    
\usepackage{hyperref}       
\usepackage{url}            
\usepackage{booktabs}       
\usepackage{amsfonts}       
\usepackage{nicefrac}       
\usepackage{microtype}      
\usepackage{lipsum}
\usepackage{graphicx}
\graphicspath{ {./images/} }
\usepackage[utf8]{inputenc} 
\usepackage[greek,english]{babel}

\title{Instruction-Tuning Pretrained Causal Language Models for Text Restoration of Ancient Greek Papyri and Inscriptions}

\author{
  Eric Cullhed \\
  Department of Linguistics and Philology\\
  Uppsala University\\
  Box 635 \\
  751 26 Uppsala \\
  Sweden \\
  \texttt{eric.cullhed@lingfil.uu.se} \\
}

\begin{document}
\maketitle
\begin{abstract}
This article presents an experiment in fine-tuning a pretrained causal language model (Meta’s Llama 3.1 8B Instruct) to assist with restoring missing or illegible characters in ancient Greek inscriptions and documentary papyri.

Utilizing a straightforward instruction-based approach and a 95\%/5\% train/test split, the papyrus restoration model achieved a character error rate (CER) of 14.9\%, a top-1 accuracy of 73.5\%, and a top-20 accuracy of 86.0\% for sequences up to 10 characters. A model was also fine-tuned for geographic attribution, reaching a top-1 accuracy of 66.4\% and a top-3 accuracy of 79.9\%. In chronological attribution, it demonstrated an average deviation of 21.7 years from the actual \emph{terminus post/ante quem}, with a median deviation of 0 years.

For inscriptions, the restoration model achieved a CER of 20.5\%, a top-1 accuracy of 63.7\%, and a top-20 accuracy of 83.0\% for sequences up to 10 characters. In geographic attribution, it attained a top-1 accuracy of 75.0\% and a top-3 accuracy of 83.7\%, while in dating, it had an average deviation of 37.1 years and a median deviation of 3 years from the actual date range.

Benchmarked against the state-of-the-art model (Ithaca) on a shared test set and on recently edited inscriptions, the instruction-tuned models excelled in text restoration, while also offering the practical advantage of ignoring spaces during reconstruction, which aligns with the \emph{scriptio continua} of ancient textual artifacts. However, their performance in geographic and chronological attribution was lower than Ithaca’s. To evaluate the approach in a more even setup, the instruction model was retrained with an 80\%/10\%/10\% train-validation-test split, and still outperformed Ithaca in text restoration.

The results are preliminary but suggest that fine-tuning larger pretrained causal language models using instruction templates for emendations and conjectures to ancient texts holds significant promise.
\end{abstract}


\section{Introduction}
A significant number of documentary texts in ancient Greek—such as announcements, laws, receipts, contracts, and letters—have survived from the Archaic period through to the early Byzantine era. These documents were inscribed or written on various materials, including blocks of stone and sheets of papyrus.

A key task of philological research is to assign dates to these texts and attribute them to their places of origin. Insights can typically be drawn from the provenance of the artifact in question, its physical features, and writing style. However, the analysis also heavily relies on clues from the textual content itself, such as the information conveyed or specific linguistic features. Another important challenge is the reconstruction of damaged or missing letters in these fragments, which are frequently marred by gaps, abrasions, and scribal errors. Grammatical and stylistic criteria, as well as parallels from similar documents—now readily searchable through databases such as \url{www.papyri.info} and \url{inscriptions.packhum.org}—are often invoked to support or challenge specific conjectures or emendations. Traditionally, such suggestions have been the product of the linguistic creativity and inventiveness of individual scholars, drawing on expertise accumulated over years of engaging with similar ancient texts. Conjectural “higher” criticism is performed \emph{ope ingenii} (“with the aid of one’s ingenuity”), as textual critics put it. It is a branch of philological research that seems to align as much with the realm of art as with science—or, at least, with a scientific field where, to borrow a phrase from T.S. Eliot \cite[p.~103]{eliot1920}, “there is no method except to be very intelligent.”

This traditional division within textual criticism—between the objective, technical analysis of manuscript data and the subjective, intuitive insight of an astute critic—is undergoing a significant transformation. Machine learning has the potential to contribute across all areas of philology, including those traditionally characterized by interpretation, conjecture, and creativity. Language models can be pretrained on the entire surviving textual record of ancient civilizations and fine-tuned to parse, classify, and analyze various aspects of words, phrases, and texts in these languages. They can be applied to explore relationships between manuscripts, identify prosodic, rhetorical, and narrative patterns, detect intertextual connections, recognize themes, classify sentiments, summarize ideas, and translate texts (see review in \cite{sommerschield-etal-2023-machine}). More directly relevant to our purposes, they can be trained to date and geographically attribute texts, detect errors, generate corrections, and fill in missing words and phrases (\cite{assael-etal-2019-restoring, assael_restoring_2022, bamman2020latinbertcontextuallanguage, cowen-breen-etal-2023-logion, 9f4bf3cfaad5415f8f09663b47919f1b}).

In the field of Greek documentary texts, Assael and colleagues \cite{assael_restoring_2022} were the leading pioneers. They trained a transformer model with a sparse attention mechanism (BigBird), combining word- and character-level tokenization, to date, attribute geographically, and reconstruct damaged sequences in ancient Greek inscriptions. On their test set of 7,811 inscriptions, they reported the following results:

\begin{itemize}
    \item \emph{Text Restoration}: The model achieved a top-1 accuracy of 61.8\% for restoring sequences of 1 to 10 characters, compared to 25.3\% for the human experts they recruited. The character error rate (CER) for the model, averaged over character lengths 1–10, was 26.3\%, while the corresponding figure for philologists was 59.6\%. Additionally, in 78.3\% of the cases, the correct reconstruction was found within the top 20 results.
    \item \emph{Geographical Attribution}: The model reached a top-1 accuracy of 70.8\%, with the correct location appearing within the top 3 in 82.1\% of cases.
    \item \emph{Chronological Attribution}: The model’s calculated dates, derived from the probability distribution over each century between 800 BCE and 800 CE, were on average 29.3 years off from the actual (\emph{termini post} and \emph{ante quem}), with a median error of only 3 years.
\end{itemize}

The aim of the following experiment is to build on this work and advance research in machine learning and textual criticism in three specific respects:

First, the model published by Assael and colleagues \cite{assael_restoring_2022} counts spaces as characters in the restoration task. For example, the sequence
\selectlanguage{greek}
και ο λογ – – – – – – ος τον θεον
\selectlanguage{english}
would not be decoded as
\selectlanguage{greek}
και ο λογος ην προς τον θεον.
\selectlanguage{english}
Instead, the input would need to be
\selectlanguage{greek}
και ο λογ – – – – – – – – ος τον θεον,
\selectlanguage{english}
requiring the researcher to guess beforehand that damaged characters 3 and 6 are empty spaces. In ancient inscriptions and papyri—typically written in \emph{scriptio continua} or at least without clearly marked spaces—we often know the approximate number of missing letters but not the precise number of missing characters and word boundaries. This study aims to develop a model that tracks only characters, better reflecting the real-world situations philologists face when working with damaged textual artifacts.

Second, given the rapid advancements in deep learning and natural language processing over the last two years, it is worth exploring how newer, larger pretrained causal language models—such as Meta’s LLaMA, Google’s Gemma, Microsoft’s Phi, or OpenAI’s GPT-4o—can assist textual critics. One promising approach could involve enabling masked language modeling in these models and retraining them to fill in missing sequences of text within a bidirectional context (see the first step of the recipe developed by \cite{behnamghader2024llm2veclargelanguagemodels}, LLM2Vec). However, in this preliminary study, I have opted for a simpler approach that is immediately applicable and open to experimentation with larger and newer models in the future: using a straightforward instruction template, prompting the model to generate the date, location, or reconstruction at the end of the sequence after a special token.

Third, I aim to expand the application of deep learning for textual criticism of ancient Greek documentary texts to include papyri in addition to inscriptions. While Pavlopoulos and colleagues \cite{pavlopoulos-etal-2023-dating} developed a regression model that dates Greek papyri with an average error of 54 years, there has otherwise been little strictly text-based deep-learning research on documentary papyri.

\section{Method}
\label{sec:headings}
Date spans, geographic attributions, and texts for the ancient Greek inscriptions were sourced from the Packard Humanities Institute database (\url{inscriptions.packhum.org/}) using the text processing pipeline developed and shared by Sommershield and colleagues \cite{sommerschield2021iphi} (\url{github.com/sommerschield/iphi}). The texts of Greek documentary papyri were obtained from the Duke Databank of Documentary Papyri (DDbDP), with their chronological and geographical metadata sourced from the Heidelberger Gesamtverzeichnis der griechischen Papyrusurkunden Ägyptens (HGV), both accessible via \url{https://papyri.info}’s GitHub repository at \url{https://github.com/papyri}.

The datasets were cleaned of diacritics, commas were removed, and all other punctuation marks were converted to a high dot, ``·’’. The data was then split into training (95\%) and test sets (5\%).

Dates for both papyri and inscriptions were formatted as discrete numbers corresponding to the midpoint of their respective date spans. For dates with only a \emph{terminus post quem}, they were set 25 years after the given date, while those with only a \emph{terminus ante quem} were set 25 years earlier. Place name attributions for the papyri were normalized semi-manually due to inconsistencies across database entries. While more detailed processing would be beneficial in the future, the current approach—though admittedly rough (see \url{https://github.com/ericu9500/PapyriAndInscriptions/train_data/data/places_and_dates.tsv})—was deemed sufficient for the proof-of-concept purposes of this study. All text editions were included in two versions: one in which philologists’ reconstructions of lost or uncertain characters were integrated, and another where these characters were replaced with ``-''. Individual missing characters were represented by a single ``-'', while longer stretches of missing text were indicated by 10 consecutive hyphens.

Text masks were generated prior to tokenization by randomly selecting spans of 3 to 20 characters, ensuring that no more than 50\% of a sequence of intact characters was masked. Longer spans of intact characters were favored by assigning weights proportional to the square of their lengths, increasing the probability of selecting longer spans for masking. These spans were replaced with a simple placeholder in the format ``[6 letters missing]’’, where the count excluded spaces and punctuation marks within the masked text segment.

The training data was formatted using the same chat template that the instruction model had been pre-tuned with, utilizing one of three system prompts: ``Date this papyrus fragment/inscription to an exact year!'', ``Assign this papyrus fragment/inscription to an exact place!'', or ``Reconstruct the missing letters in this papyrus fragment/inscription!''. The user input consisted of a version of the text, while the assistant’s response provided the corresponding date, place, or masked sequence of characters, including spaces and punctuation marks. The text was tokenized using the model’s TikToken-based tokenizer.

The base model, LLaMA 3.1 8B-Instruct \cite{dubey2024llama3herdmodels}, was fine-tuned on 4 A100 GPUs with 40 GB of memory and a batch size of 8, or on 80 GB GPUs with a batch size of 20, depending on availability at the time, using the toolkit Torchtune (\url{github.com/pytorch/torchtune}). In an initial calibration round, exploratory fine-tuning was conducted with three different data combinations to identify the best approach: one model trained solely on papyri (4 epochs, approximately 80 hours), another solely on inscriptions (4 epochs, approximately 120 hours), and a third on both inscriptions and papyri (3 epochs, approximately 132 hours). For each data combination, two versions were trained: one using the entire sequence of system, user, and assistant messages, and another with the prompt masked out for loss computation.

Subsequently, the best-performing models were further trained separately for the three tasks until their performance plateaued on the test set. During this stage, entries shorter than 75 tokens or longer than 847 tokens were filtered out, and data augmentation was applied by randomly altering the order of sentences and adding noise through the replacement of 5\%, 10\%, 15\%, 20\%, and 25\% of preserved characters with ``-''. Additionally, masked sequences of 1–20 characters (not only 3–20) were included. In the final stage, the fine-tuned models were re-merged with the base model using the TIES method developed by Yadav and colleagues \cite{yadav2023tiesmergingresolvinginterferencemerging}, utilizing the Mergekit toolkit (\url{github.com/arcee-ai/mergekit}).

The models were evaluated using the same metrics proposed by Assael and colleagues \cite{assael_restoring_2022}: CER (Character Error Rate) averaged across lengths 1–10, top-1 and top-20 accuracy for normalized reconstructions (where spaces, punctuation marks, and numerals were ignored), top-1 and top-3 accuracy for geographical attributions, and average and median deviation from the date ranges provided in the database. Entries shorter than 90 characters were excluded. For inscriptions, the test was conducted on 7,811 samples for restoration, 2,870 samples for geographical attribution, and 1,856 samples for dating. For papyri, the test was conducted on 7,811 samples for restoration, 1,990 samples for geographical attribution, and 2,295 samples for dating. Dates, places, and reconstructions were evaluated based on top-1, top-3, and top-20 rankings, respectively, out of 60 beams.

In an additional step, the models were fine-tuned for restoration on texts truncated to 50–750 characters to allow for comparison with the model developed by Assael and colleagues \cite{assael_restoring_2022}. They were tested on a subset of my test set that coincided with theirs (entries with a PHI ID ending in 3). This subset included 264 samples with dates, 435 samples with places, and 4,110 masked samples based on 411 papyri for restoration (100 samples of each character length). Additionally, the models were evaluated on 2 previously unedited Uppsala papyri (1,620 masked samples) and 6 recently edited inscriptions (4,111 masked samples) to assess their performance on texts that could not have been included in the pretraining data of Meta’s base model. Subpages of both \url{http://papyri.info} and \url{https://inscriptions.packhum.org} sporadically appear in the indices of the Common Crawl corpus (\url{http://index.commoncrawl.org}). In a final step, the inscription restoration model was fine-tuned from scratch based on the papyrus restoration model using the same 80\%/10\%/10\% train-validation-test split as Assael and colleagues \cite{assael_restoring_2022}, augmenting the training data with OpenAI's GPT-4o-mini (see the dataset on \url{https://huggingface.co/datasets/Ericu950/Inscriptions_2}).

\section{Results}\label{sec3}
\subsection{Calibration}\label{subsec1}
For the reconstruction task, training with the prompt masked out was clearly superior, resulting in a difference of 8\% CER for the inscriptions and 7\% for the papyri after 4 epochs, all other factors being equal. The transfer effect of training the same model on both papyri and inscriptions was slightly negative, with a 5\% difference for inscriptions and 0.1\% difference for papyri after 3 epochs.

For the geographical attribution task, training with the prompt masked out was also superior, yielding a 10.8\% difference in top-1 accuracy for inscriptions and a 4.9\% difference for papyri after 4 epochs. The transfer effect of training the same model on both papyri and inscriptions was negative, with a significant 39.2\% difference for inscriptions and a smaller 1.8\% difference for papyri after 3 epochs.

For the dating task, again, training with the prompt masked out led to better performance, with an average difference of 16 years for inscriptions and 3 years for papyri after 4 epochs. The transfer effect of training the same model on both papyri and inscriptions was slightly positive for papyri (-2 years) but negative for inscriptions (0.3 years) after 3 epochs.

The top models in this round were as follows:

\begin{itemize}
    \item \emph{Reconstruction}: Three epochs for inscriptions (CER 27.9\%, top-1 accuracy 52.2\%, and top-20 accuracy 71.0\%) and four epochs for papyri (CER 20.6\%, top-1 accuracy 64.0\%, and top-20 accuracy 80.86\%).
    \item \emph{Geographical Attribution}: Three epochs for inscriptions (top-1 accuracy 71.0\%, top-3 accuracy 81.4\%) and four epochs for papyri (top-1 accuracy 62.3\%, top-3 accuracy 77.9\%).
    \item \emph{Dating}: Three epochs for inscriptions (average distance 40.3 years, median distance 5 years) and three epochs (trained on both papyri and inscriptions) for papyri (average distance 22.11 years, median distance 0 years).
\end{itemize}

\subsubsection{Continued training}\label{subsubsec2}
The calibration round provided several key insights: it is beneficial to mask out the prompt; the effects of transfer learning between papyri and inscriptions were slight or negative; chronological and geographical attributions plateaued quicker than reconstructions; and there was early deterioration in reconstruction performance, highlighting the need for data augmentation techniques. As a result, three separate models were trained for each task and each corpus. Due to slower progress with the inscriptions dataset, characterized by more short entries, texts corresponding to less than 75 or more than 847 tokens were filtered out. Data augmentation techniques, as detailed in the method section above, were applied. Additionally, the decision to mask 3–20 characters instead of 1–20 resulted in long and rambling reconstructions when the test set required the reconstruction of 1–2 characters. Therefore, samples with 1–2 characters masked were also included in this phase of training.

For the reconstruction task, performance plateaued after 6 additional epochs (for a total of 9 epochs) for the inscriptions, achieving a CER of 22.3\%, top-1 accuracy of 61.0\%, and top-20 accuracy of 74.2\%. Training on the papyrus dataset plateaued after 7 additional epochs (for a total of 11 epochs), with a CER of 16.3\%, top-1 accuracy of 69.9\%, and top-20 accuracy of 80.7\%.

For the geographical attribution task, performance plateaued after 4 additional epochs (for a total of 7 epochs) for the inscriptions, achieving a top-1 accuracy of 75.0\% and top-3 accuracy of 83.0\%. Training on the papyrus dataset for geographical attribution plateaued after 5 additional epochs (for a total of 9 epochs), with a top-1 accuracy of 66.4\% and top-3 accuracy of 79.9\%.

For the dating task, performance plateaued after 1 additional epoch (for a total of 4 epochs) for the inscriptions, with an average deviation of 37.1 years and a median deviation of 3 years from the actual dating span. Training on the papyrus dataset for dating also plateaued after 1 additional epoch (for a total of 4 epochs), with an average deviation of 21.7 years and a median deviation of 0 years from the actual dating span.

\subsubsection{Re-merging}\label{subsubsec3}
In the final step, the effects of re-merging the fine-tuned models with the original base model were assessed. The TrIm, Elect Sign \& Merge method (TIES-Merging), where parameters that changed only slightly during fine-tuning are trimmed (\cite{yadav2023tiesmergingresolvinginterferencemerging}), was applied, using the toolkit Mergekit (\url{github.com/arcee-ai/mergekit}). This process led to a deterioration of 10\% in geographical attribution accuracy for inscriptions and 5\% for papyri, along with an increase in the average error for dating by 4 years for inscriptions and 3 years for papyri. For the reconstruction tasks, however, there was only insignificant deterioration of 0.1–0.2\% in CER and top-1 accuracy, but a 3\% increase in top-20 accuracy. Hence, the model’s overall restoration capacity interestingly improved when it was re-infused with the base model.

\subsubsection{Full comparison with Ithaca}\label{subsubsec4}
An accurate comparison with the state-of-the-art model for these tasks developed by Assael and colleagues \cite{assael_restoring_2022} – Ithaca – was at first not possible, since the present model was trained for experimental purposes with a 95\%/5\% split as opposed to Ithaca's 80\%/10\%/10\%. Yet some form of comparison was considered useful. Accordingly, in a penultimate step, the models were fine-tuned on the dataset where every entry was truncated to 50–750 characters (minimum and maximum length of entries for Ithaca). The resulting inscription restoration model and Ithaca were then run on two test sets: one solely derived from entries within my 5\% test set that coincided with Assael and colleagues' test split (where PHI ID ends in 3), and one derived from inscriptions that were not in the PHI or openly available on the internet as of December 2023 (4111 samples based on 6 inscriptions; see appendix). On the shared test set, the fine-tuned Llama model had an average CER of 22.55\% (vs. Ithaca's 27.0\%), a top-1 accuracy of 61.4\% (vs. 60.9\%), and top-20 accuracy of 77.00\% (vs. 71.6\%). On the new inscriptions, it had a CER of 30.8\% (vs. Ithaca's 31.1\%), a top-1 accuracy of 48.9\% (vs. 46.5\%), and a top-20 accuracy of 72.9\% (vs. 68.3\%).

In geographic attribution of inscriptions in the shared test set, the Llama model had a top-1 accuracy of 65.0\% (vs. Ithaca's 69.0\%) and top-20 accuracy of 78.4\% (vs. Ithaca's 80.2\%). In chronological attribution, it had an average distance of 55.7 (vs. Ithaca's 48.3) and a median distance of 4 (vs. Ithaca's 4.5).

The Papyrus model on the recent papyri showed variability: in the case of Pap.Ups 106, it had a CER of 21.0\%, top-1 accuracy of 62.5\%, and top-20 accuracy of 79\%. With Pap.Ups. 18, it had a CER of 38.6\%, top-1 accuracy of 37.2\%, and top-20 accuracy of 61.2\%. It plausibly assigns the former document to Oxyrhynchus and accurately dates it to 71 AD (the fourth year of Vespasian, 71–72 AD); equally plausibly, it places the latter in Oxyrhynchus and dates it to 304 AD, approximately 11–12 years from the correct date of around 292 AD (the eighth year of Diocletian and the seventh of Maximian are mentioned).

These further-tuned models were also tested on the entire 5\% test set with good results. The inscription model had a final CER of 20.5\%, a top-1 accuracy of 63.7\%, and top-20 accuracy of 83.0\%. The Papyrus model had a CER of 14.9\%, top-1 accuracy of 73.5\%, and top-20 accuracy of 85.9\%.

In order to conduct a more exact comparison between the approach explored in this paper and Ithaca, an attempt to use the same 80\%/10\%/10\% split was made. The papyrus restoration model was used as the base model. Only inscriptions with an ID \emph{not} ending in 3 or 4 were used. For data augmentation, each inscription was processed with OpenAI's GPT-4o mini to create a modified version of the text with 10 slight variations of the following system prompt:

\begin{quote}
Generate a modified version of this inscription for training a model to restore gaps in ancient inscriptions. Apply advanced data augmentation techniques, such as reordering words and sentences, and replacing as many words and phrases as possible with their synonyms, while strictly preserving Greek morphology, syntax, proper names, and place names. Ensure creativity in restructuring, but avoid altering the essential meaning. Only provide the altered inscription, without additional commentary.
\end{quote}

Performance on the validation set plateaued after 8 epochs. On the exact same subsection of the data used in the previous test, it achieved a somewhat higher yet competitive CER of 21.8\%, top-1 accuracy of 62.3\%, and top-20 accuracy of 78.0\%. On the test set of recently edited inscriptions, it outperformed the previous model with a CER of 28.6\%, a top-1 accuracy of 51.3\%, and top-20 accuracy of 68.6\%.

\section{Discussion}\label{sec4}
\begin{table*}[t]
\caption{Results for restoration, geographical attribution, and dating. \label{tab:results_comparison}}
\centering
\scriptsize 
\tabcolsep=5pt 
\begin{tabular*}{\textwidth}{@{\extracolsep{\fill}}lcccccc@{\extracolsep{\fill}}}
\toprule
& \multicolumn{3}{c}{Restoration} & \multicolumn{2}{c}{Region} & \multicolumn{1}{c}{Date} 
\rule{0pt}{15pt}
\\
\cline{2-4}\cline{5-6}\cline{7-7}
\rule{0pt}{10pt} 
& CER (\%) & Top-1 (\%) & Top-20 (\%) & Top-1 (\%) & Top-3 (\%) & \multicolumn{1}{c}{Years} \\
\midrule
Epigr\_1/2\_Llama3.1-8b-Instruct on my test & 20.5 & 63.7 & 83.0 & 75.0 & 83.0 & 37.1 \\
Assael et al. (2022) on their test & 26.3 & 61.8 & 78.3 & 70.8 & 82.1 & 29.3 \\
\midrule
Epigr\_3\_Llama3.1-8b-Instruct on shared test & \textbf{21.8} & \textbf{62.3} & \textbf{78.0} & 65.0 & 78.4 & 55.7 \\
Assael et al. (2022) on shared test set & 27.0 & 60.9 & 71.6 & \textbf{69.0} & \textbf{80.2} & \textbf{48.3} \\
\midrule
Epigr\_3\_Llama3.1-8b-Instruct on new inscr. & \textbf{28.6} & \textbf{51.3} & \textbf{68.6} & - & - & - \\
Assael et al. (2022) on new inscr. & 31.1 & 46.5 & 68.3 & - & - & - \\
\midrule
Papy\_1\_Llama3.1-8b-Instruct on my test & 14.9 & 73.5 & 85.9 & 66.4 & 79.9 & 21.7 \\
Papy\_1\_Llama3.1-8b-Instruct on two new papyri & 21.0/38.6 & 62.5/37.2 & 79.0/61.2 & - & - & - \\
Pavlopoulos et al. (2023) on their test & - & - & - & - & - & 54 \\
\bottomrule
\end{tabular*}
\end{table*}

Table \ref{tab:results_comparison} summarizes the results, providing a direct comparison between the models developed in this study and Ithaca \cite{assael_restoring_2022}. The superior results on the shared test set as well as the test set with unedited or recently edited papyri and inscriptions suggest that the strength of these models does not simply stem from possible exposure to test texts included in the data used during Meta’s pretraining process, but rather from the effects of the fine-tuning process itself. Although they exhibit a higher CER and lower top-1 and top-20 accuracy on this material, the fact that Ithaca's performance was similarly affected, along with the variation in results between the two different papyri, suggests that the challenge lies in the task itself rather than a lack of prior exposure.

In any case, the primary goal of this experiment was to assess whether a simple, instruction-based approach could compete with more established methods in this area, and the findings indicate that it can. These results justify further exploration through additional experiments involving smaller, larger, and more advanced base models, as well as alternative strategies for dataset formatting, masking, training, and model merging. The tools developed in this study outperform the state-of-the-art in text restoration, demonstrating significant potential. However, the method’s effectiveness for dating and geographic attribution was less conclusive.

Moreover, these models offer a notable advantage: they are particularly useful in cases where scholars have an approximate sense of the number of missing characters but lack precise information about missing word boundaries—a frequent challenge in the study of ancient textual artifacts. Furthermore, the study proved the effectiveness of using large language models for data augmentation to improve text restoration models for Ancient Greek texts. Additionally, the papyri models introduced here represent a completely new resource for this material.

\section{Conclusion}\label{sec5}
This experiment demonstrated the efficacy of fine-tuning pretrained causal language models for text restoration in ancient Greek inscriptions and documentary papyri using a simple instruction template, surpassing the current state of the art. The simplicity and scalability of this method, particularly the ease of fine-tuning newer models on the same data with minimal modification, is a key advantage. There is significant potential for further improvement through refined data cleaning, data augmentation, and masking strategies, optimized training parameters, and experimentation with smaller, larger, and more advanced base models as they become available. However, the method was less successful than previous models for dating and geographic attribution.

Although more specialized models may eventually outperform this straightforward method, instruction-tuned pretrained causal language models can serve as an efficient and easily implementable moving baseline. Researchers developing machine learning methods for textual criticism could aim either to surpass this baseline or to incorporate it as part of a more comprehensive solution. Rather than viewing different models as competitors, a more productive approach might involve combining them as nodes in a system supported by an adjudicator or ranker model. This collaborative approach, drawing on the complementary strengths of various AI models, likely holds the greatest promise for developing powerful tools to support philological research.

\bibliographystyle{unsrt}  
\bibliography{references}

\begin{thebibliography}{10}

\bibitem{eliot1920}
T.S. Eliot.
\newblock The perfect critic ii.
\newblock {\em Athenaeum}, pages 102--104, 1920.

\bibitem{sommerschield-etal-2023-machine}
Thea Sommerschield, Yannis Assael, John Pavlopoulos, Vanessa Stefanak, Andrew Senior, Chris Dyer, John Bodel, Jonathan Prag, Ion Androutsopoulos, and Nando de~Freitas.
\newblock Machine learning for ancient languages: A survey.
\newblock {\em Computational Linguistics}, pages 703--747, September 2023.

\bibitem{assael-etal-2019-restoring}
Yannis Assael, Thea Sommerschield, and Jonathan Prag.
\newblock Restoring ancient text using deep learning: a case study on {G}reek epigraphy.
\newblock In Kentaro Inui, Jing Jiang, Vincent Ng, and Xiaojun Wan, editors, {\em Proceedings of the 2019 Conference on Empirical Methods in Natural Language Processing and the 9th International Joint Conference on Natural Language Processing (EMNLP-IJCNLP)}, pages 6368--6375, Hong Kong, China, November 2019. Association for Computational Linguistics.

\bibitem{assael_restoring_2022}
Yannis Assael, Thea Sommerschield, Brendan Shillingford, Mahyar Bordbar, John Pavlopoulos, Marita Chatzipanagiotou, Ion Androutsopoulos, Jonathan Prag, and Nando de~Freitas.
\newblock Restoring and attributing ancient texts using deep neural networks.
\newblock {\em Nature}, 603(7900):280--283, March 2022.
\newblock Number: 7900 Publisher: Nature Publishing Group.

\bibitem{bamman2020latinbertcontextuallanguage}
David Bamman and Patrick~J. Burns.
\newblock Latin bert: A contextual language model for classical philology, 2020.

\bibitem{cowen-breen-etal-2023-logion}
Charlie Cowen-Breen, Creston Brooks, Barbara Graziosi, and Johannes Haubold.
\newblock Logion: Machine-learning based detection and correction of textual errors in {G}reek philology.
\newblock In Adam Anderson, Shai Gordin, Bin Li, Yudong Liu, and Marco~C. Passarotti, editors, {\em Proceedings of the Ancient Language Processing Workshop}, pages 170--178, Varna, Bulgaria, September 2023. INCOMA Ltd., Shoumen, Bulgaria.

\bibitem{9f4bf3cfaad5415f8f09663b47919f1b}
Barbara Graziosi, Johannes Haubold, Charlie Cowen-Breen, and Creston Brooks.
\newblock Machine learning and the future of philology: A case study.
\newblock {\em TAPA}, 153(1):253--284, March 2023.
\newblock Publisher Copyright: {\textcopyright} 2023 by the Society for Classical Studies.

\bibitem{behnamghader2024llm2veclargelanguagemodels}
Parishad BehnamGhader, Vaibhav Adlakha, Marius Mosbach, Dzmitry Bahdanau, Nicolas Chapados, and Siva Reddy.
\newblock Llm2vec: Large language models are secretly powerful text encoders, 2024.

\bibitem{pavlopoulos-etal-2023-dating}
John Pavlopoulos, Maria Konstantinidou, Isabelle Marthot-Santaniello, Holger Essler, and Asimina Paparigopoulou.
\newblock Dating {G}reek papyri with text regression.
\newblock In Anna Rogers, Jordan Boyd-Graber, and Naoaki Okazaki, editors, {\em Proceedings of the 61st Annual Meeting of the Association for Computational Linguistics (Volume 1: Long Papers)}, pages 10001--10013, Toronto, Canada, July 2023. Association for Computational Linguistics.

\bibitem{sommerschield2021iphi}
Thea Sommerschield*, Yannis Assael*, Brendan Shillingford, Mahyar Bordbar, John Pavlopoulos, Marita Chatzipanagiotou, Ion Androutsopoulos, Jonathan Prag, and Nando de~Freitas.
\newblock {I.PHI} dataset: ancient greek inscriptions.
\newblock \url{https://github.com/sommerschield/iphi}, 2021.

\bibitem{dubey2024llama3herdmodels}
Abhimanyu Dubey et~al.
\newblock The llama 3 herd of models, 2024.

\bibitem{yadav2023tiesmergingresolvinginterferencemerging}
Prateek Yadav, Derek Tam, Leshem Choshen, Colin Raffel, and Mohit Bansal.
\newblock Ties-merging: Resolving interference when merging models, 2023.

\bibitem{akinci2024new}
E.~Akinci {\"O}zt{\"u}rk, M.~Ricl, and C.~Tanriver.
\newblock A new confession inscription from the sanctuary of apollo lairmenos.
\newblock {\em Zeitschrift f{\"u}r Papyrologie und Epigraphik}, 230:62--66, 2024.

\bibitem{blumell2024christian}
Lincoln~H. Blumell and Amer~El Mesiry.
\newblock Some christian epitaphs in greek in the magazine at fustât (old cairo).
\newblock {\em Zeitschrift f{\"u}r Papyrologie und Epigraphik}, 230:153--166, 2024.

\bibitem{staab2023bauinschrift}
Gregor Staab.
\newblock Bauinschrift in iambischen trimetern anlässlich des wiederaufbaus einer stoa in elusa durch den kaiserlichen arzt stephanos (ca. 340–360 n. chr.).
\newblock {\em Zeitschrift f{\"u}r Papyrologie und Epigraphik}, 228:89--96, 2023.

\bibitem{liddel2023new}
Peter Liddel.
\newblock New greek inscriptions in uk collections part ii: Four new greek inscriptions at the british museum.
\newblock {\em Zeitschrift f{\"u}r Papyrologie und Epigraphik}, 225:159--168, 2023.

\bibitem{camia2023epitaph}
Francesco Camia and Elisa~Nuria Merisio.
\newblock Un epitaffio metrico cristiano da elaiussa sebaste (cilicia).
\newblock {\em Zeitschrift f{\"u}r Papyrologie und Epigraphik}, 226:60--64, 2023.

\end{thebibliography}

\section{Appendix: Preliminary transcriptions of two unedited Uppsala papyri and texts of 6 recently edited inscriptions that were used}
\subsection{Pap.Ups. 106}
\selectlanguage{greek}
ετους τεταρτου αυτοκρατορος καισαρος ουεσπασιανου σεβαστου - - - - - - - - - - - - - - - - - - ομολογει παυσιριων απολλωνιου του παυσιριωνος μητρος - - - - - - - - - - - - - - -τωι γεγονοτι αυτωι εκ της γενομενης και μετηλλαχυιας αυτου γυναικος - - - - - - - - - - - - - - - - - - - - - - - - - απο της αυτης πολεως εν αγυιαι συγχωρειν ειναι - - - - - - - - - - - - - - - - - - - - - - - - - - - - - - - - - - - - - - - - - - - - - - - - - - - - - -ς αυτωι εξ ης συνεστιν - - - - - - - - - - - - - - - - - - - - - - - - - - - - - - - - - - - - - - - -της αυτης γενεας την υπαρχουσαν αυτωι οικιαν - - - - - - - - - - - - - - - - - - - - - - - - - - - - - - - - - - - - - - -καὶ αιθριον και αυλη απερ ο υιος διοκορος - - - - - - - - - - - - - - - - - - - - - - - - - - - - - - - - - -εγραψεν του δ αυτου διοσκορου ειναι - - - - - - - - - - - - - - - - - - - - - - - - - - - - - - - - - - - - - - - - - - - - - - και προ κατενγεγυηται τα δικαια - - - - - - - - - - - - - - - - - - - - - - - - - - - - - - - - - - - - - - νης κατα τους της χωρας νομουσ· εαν δε μη - - - - - - - - - - - - - - - - - - - - - - - - - - - - - - - - - - - - - - - υπ αυτου τηι του διοσκορου σημαινομενηι - - - - - - - - - - - - - - - - - - - - - - - - - - - - - - - - - - -ενοικισμωι του ημισους μερους της προκειμενης οικιας - - - - - - - - - - - - - - - - - - - - - - - - - - - - - - - - - διοσκορος την τουτων αποχην - - - - - - - - - - - - - - - - - - - - - - - - - - - - - - - - - - - - - - - - - - - - -μηδ υπεναντιον τουτοις επιτελειν μηδε - - - - - - - - - - - - - - - - - - - - - - - - - - - - - - - - - - - - - - - - - - - - - - - - ανασκευηι κατ αυτης τιθεσθαι ομολογιαν μηδε - - - - - - - - - - - - - - - - - - - - - - - - - - - - - - - - - - - επιτελεσαι η χωρις του κυρια ειναι τα διομολογημενα παραβαινειν, εκτεινειν δε τον παραβησομενον τωι υιωι διοσκορωι η τοις παρ αυτου καθ εκαστην εφοδον το τε βλαβος και επιτιμον αργυριου δραχμας 0 και εις το δημοσιον τας ισας και μηθεν ησσον· δ - - - - -ιων ομολογιαν συνεχωρησεν· 
\selectlanguage{english}
\\ I am grateful to Disa Lindholm for her previous transcription of this text. 

\subsection{Pap.Ups. 18}
\selectlanguage{greek}
λογιστηι· απο της - - - - - - - ων λαμπρας πολεως παρα α - - - - - - - πακτουμηιας διονυσοδωρας - - - - - - - - - - - - - - - απο της αυτης πολεωσ. ενετυχον επι υπομνηματων τω κυριω ημων τω διασημοτατω ηγεμονι σατριω αρριανω δια φιλεου αποσυσταθεντος υπ εμου - - - - - ενδειξας τω μεγαλειω αυτου ως πρωην τετελευτηκοτος του ενος των αφ- - - - - - - - - - - - - - - - - -ς πρεσβυτερου-αιτου. υπο δε ετερου αυτου αδελφου ηρωνος τ- - - - - - - - - - - - - - - - -τος εν τω δεσμωτηριω αυτην- - - - - -ην- - - - - - - - - - - -κληρονομιας του κατοιχομενου υιου μου ηνπερ αποσπασαι εια- - - - - - - - - - - - - - -νομους ε- - - - - - - -καιος ατιλιος αρτεμιδωρος εφρασ- - - - - - - - - - - - - - - - - - - - - - - - -νει τοινυν- - - - - - -ην απ’ αυτου το μεν δη- - -μηδεν βιαιον α- - - - - - - - - - - - - - - - -προνοιας π- - - - - -κω προ δικης προχω- - - -αναγκαιως το με- - - - - - - - - - - - - -γενηματων- - - - - - - -σφαλεραν καθιστη ω- - - - -επιμελεια αξιουσα- - - - - - - - - - - -αντιδικου κ- - - - - - - - -προς το ειναι με εν- - - - - - -και- - - - - - -των- - - - - - - - - - - -χρηματων κατ- - - - - -φαντα τη ηγεμονια κ- - - - - - - - - - - - - - - - - - - - - - - - - - - - - -ου και ετουσ- - - -των κυριων ημων διοκλητιανου και μαξιμιανου σεβαστιων και ετους ια των κυριων ημων κωνσταντιου και μαξιμιανου- - - - - - - - - - - - -των καισαρων παχων- - - - -αντιγραφον του υπομνηματοσ- - - - - - - - - -των κυριων ημων διοκλητιανου το η και μαξιμιανου το ζ σεβαστων- - - - - - - - - - - - - - - -εν τω σηκρητω- - -παχων ε- - - - - - - - - - - - - - -ος ο διασημοτατος ηγεμων ε- - - - - - - - - - -διονυσοδωρα κατα- - - - - - - - - - - - - - -προ δικησ- - - -ι υποστησεται- - - - - - - -δια του- - -στου- - - - - - - - - - - - - - - - -κν·
\selectlanguage{english}
\\ I am grateful to Disa Lindholm for her previous transcription of this text.
\subsection{Confession Inscription from Asartepe}
\selectlanguage{greek}
σωφρων ημαρτεις ποσον επι το χωριον του λαιρμηνου και προς γυναικα και το ορος το ιερον εκοψα και βοσχησα και αλλην γυναικα φιλησας ημαρτην προς αυτην επι το χωριον και εν παλληκιω πεινων αναποθυτον εφαγα και κολασθεις ετη δυο κατα των ανανκαιων και πολλα εξομολογησαμενος εστηλλογραφησα διο παραγγελλω μηδενα καταφρονειν ηλιω απολλωνι λαιρμηνω επει εξει την εμην στηλλην εξενπλαριον·
\selectlanguage{english}
\\Edited by \cite{akinci2024new}.

\subsection{Epitaph from Fustât}
\selectlanguage{greek}
εκοιμηθη ο μακαριος βικτορος εν μηνι φαμενωθ ογδοη ινδιονος ενατη ετους διοκλητιανου τϙζ ο θεε του αγιου προδρομου και βαπτιστου ιωαννου και παντος του χορου των αγιων μαρτυρων αναπαυσον την ψυχην αυτου μετα παντων·
\selectlanguage{english}
\\Edited by \cite{blumell2024christian}.

\subsection{Building inscription from Elusa}
\selectlanguage{greek}
ταυτην θεασαι δευτεραν εκτισμενην στοαν απ ακρου περιφερους υψουμενην κτισεν ο χορηγος χρυσιου ο προστατης αρχων ιατρων στεφανος ος του δεσποτου το σωμα θεραπευει επιστατουσι δε εργοισιν ειρηναιος ευδαιμων θ αμα λαμπρος μεν ουτος ο δε νομους ησκημενος·
\selectlanguage{english}
\\Edited by \cite{staab2023bauinschrift}.

\subsection{Inscribed altar from Egypt}
\selectlanguage{greek}
ηρακλει επηκοω απολλωνιος μυτιληναιος λατομος χαριστηριον·
\selectlanguage{english}
\\Edited by \cite{liddel2023new}.

\subsection{Funerary monument from Propontis}
\selectlanguage{greek}
πριμος και ρωμανος χερσονησου και σφηκλας υιοι και ρωμανος και πριμος φροντωνος και παυλας υιοι δημητριω παππω ιδιω μνημης χαριν ονησιμος νεγωτιατωρ δημητριω ιδιω μητρωνι χρυσουν στεφανον·
\selectlanguage{english}
\\Edited by \cite{liddel2023new}.

\subsection{Christian Epitaph from Elaiussa Sebaste}
\selectlanguage{greek}
- - - - - - επτα πλησαμενον λυκαβαντας αεθλητηρα θεοιο βασιλιον ταφος ουτος ον ου χαδε γαια φυλασσει·
\selectlanguage{english}
\\Edited by \cite{camia2023epitaph}.

\subsection{Code availability}
The scripts for formatting training and test data, configuration files for training and merging, and notebooks for testing the models and calculating scores based on pre-made test outputs are available at \url{https://github.com/ericu9500/PapyriAndInscriptions}. The datasets and top-performing models are available at \url{https://huggingface.co/collections/Ericu950/papyri-and-inscriptions-66ed3af86b665725dcc28ca5}.

\section{Competing Interests}
The author declares no competing interests.

\section{Acknowledgements}
 I wish to express my gratitude to Yannis Assael and Thea Sommerschield for very helpful comments on an earlier draft. As always, the remaining errors are my own.

\section{Funding}
The computations and storage during training were enabled by resources provided by the National Academic Infrastructure for Supercomputing in Sweden (NAISS), partially funded by the Swedish Research Council through grant agreement no. 2022-06725.

\end{document}